\documentclass[letterpaper, 10 pt, journal, twoside]{IEEEtran}  
\usepackage{pgfplots}
\usetikzlibrary{pgfplots.groupplots}

\usepackage{tabularx}
\usepackage{graphicx}

\usepackage{caption}
\usepackage{subcaption}

\usepackage{amsmath}
\usepackage{amssymb}

\usepackage{multirow}
\usepackage{booktabs}

\usepackage{etoolbox}
\usepackage{cuted} 

\usepackage{url}

\usepackage{siunitx}
\DeclareSIUnit{\rad}{rad}

\usepackage{xfrac}

\usepackage[shortcuts, acronym]{glossaries}
\newacronym{icp}{ICP}{Iterative Closest Points}
\newacronym{lidar}{lidar}{Light Detection and Ranging}
\newacronym{imu}{IMU}{Inertial Measurement Unit}
\newacronym{gnss}{GNSS}{Global Navigation Satellite System}
\newacronym{kiss}{KISS-ICP}{KISS-ICP}
\newacronym{osm}{OSM}{OpenStreetMap}
\newacronym{ate}{ATE}{Absolute Translational Error}
\newacronym{nn}{NN}{Nearest Neighbour}
\newacronym{slam}{SLAM}{Simultaneous Localization and Mapping}
\newacronym{pf}{PF}{Particle Filter}
\newacronym{mcl}{MCL}{Monte Carlo Localization}
\newacronym{cnn}{CNN}{Convolutional Neural Network}
\newacronym{gicp}{GICP}{Generalized Iterative Closest Point}
\newacronym{ins}{INS}{Inertial Navigation System}
\newacronym{pps}{PPS}{Pulse Per Second}
\newacronym{rpe}{RPE}{Relative Position Error}
\newacronym{sam}{SAM}{Smoothing and Mapping}
\newacronym{lo}{LO}{Lidar Odometry}
\newacronym{lio}{LIO}{Lidar Inertial Odometry}
\newacronym{rte}{RTE}{Relative Translational Error}
\newacronym{ndt}{NDT}{Normal Distribution Transform}
\newacronym{dof}{DOF}{degree of freedom}

\definecolor{grey}{RGB}{176,176,176}
\definecolor{C0}{RGB}{31,119,180}
\definecolor{C1}{RGB}{255,127,14}
\definecolor{C2}{RGB}{44,160,44}
\definecolor{C3}{RGB}{176,176,176}
\definecolor{C4}{RGB}{148,103,189}
\definecolor{C5}{RGB}{140,86,75}
\definecolor{C6}{RGB}{227,119,194}
\definecolor{C7}{RGB}{127,127,127}
\definecolor{C8}{RGB}{188,189,34}
\definecolor{C9}{RGB}{23,190,207}

\newcommand\T{\mathbf{T}}
\newcommand\SE[1]{\text{SE}(#1)}

\newcommand\set[1]{\mathcal{#1}}

\newcommand\state{\mathbf{x}}
\newcommand\measurement{\mathbf{z}}
\newcommand\meas{\measurement}

\newcommand\information{\mathbf{\Omega}}

\renewcommand\vec[1]{\mathbf{#1}}

\begin{document}
\title{\LARGE \bf
Online Lidar-Only Odometry with Retrospective Refinement \\ of Overlapping Submaps
}
\author{Aaron Kurda, Simon Steuernagel, and Marcus Baum
    \thanks{
    This work was funded by the German Federal Ministry for Economic Affairs and Climate Action (BMWK) within the research project “OKULAr” (Grant No. 19A22003C).%
    }
    \thanks{
    The authors are with the Institute of Computer Science, University of Goettingen, Goettingen, Germany. 
    \newline E-mail: 
    {\tt\small \{aaron.kurda, simon.steuernagel, marcus.baum\}@cs.uni-goettingen.de}
    }
}

\makeatother

\maketitle

\begin{abstract}
\IEEEPARstart{L}{idar} odometry aims to estimate the ego-motion of a mobile platform from a stream of lidar scans. Traditional scan-to-map approaches register each scan against a single, evolving map, which propagates registration errors over time. To mitigate this, we propose a multi-submap approach where the current scan is registered against multiple overlapping submaps instead of a single static map. By optimizing the resulting constraints in a pose graph, our method enables not only precise estimation of the current pose, but also retrospective refinement of the submaps' anchor points, which improves short-term consistency and long-term accuracy. We demonstrate that our approach achieves significant accuracy gains over several state-of-the-art lidar odometry methods on a variety of automotive datasets while simultaneously maintaining real-time performance. Ablation studies confirm the critical role of multiple registrations and retrospective refinement of the map as core factors for our accuracy gains. Code and raw results are available on our public GitHub at~\url{https://github.com/Fusion-Goettingen/RAL_2026_Kurda_Online}.
\end{abstract}

\section{Introduction}

Odometry is the task of estimating the movement of a mobile platform with respect to a given starting position and is therefore a prerequisite for many other tasks. 
The ability of lidar sensors to sense their environment make them particularly interesting as their data can be also used for mapping~\cite{mohamed2019survey}. This makes lidar odometry especially suitable for robots that aim at navigating space autonomously. The general idea behind lidar odometry is to infer the motion of a mobile platform relative to its (static) environment by aligning the measurements of the sensor over a short period of time. A common method for this alignment is the~\ac{icp}~\cite{besl1992method} algorithm, which takes two point clouds, i.e., lidar scans, and finds the transformation that best aligns them. This process is also called registration and its solution provides an estimate of the motion between the two recordings. To infer the motion over longer periods, a simple solution is to recursively apply the registration method to consecutive pairs of lidar scans~\cite{pomerleau2013comparing}. The resulting list of motions then describe the traversed path of the agent. The recursive estimation of the pose from the motion is also called dead reckoning and is widely used in inertial odometry using \acp{imu} sensors. Its key problem is that the errors of individual registrations accumulate over time, resulting in a drifting estimate of the traversed path.

\begin{figure}
    \includegraphics[width=\columnwidth]{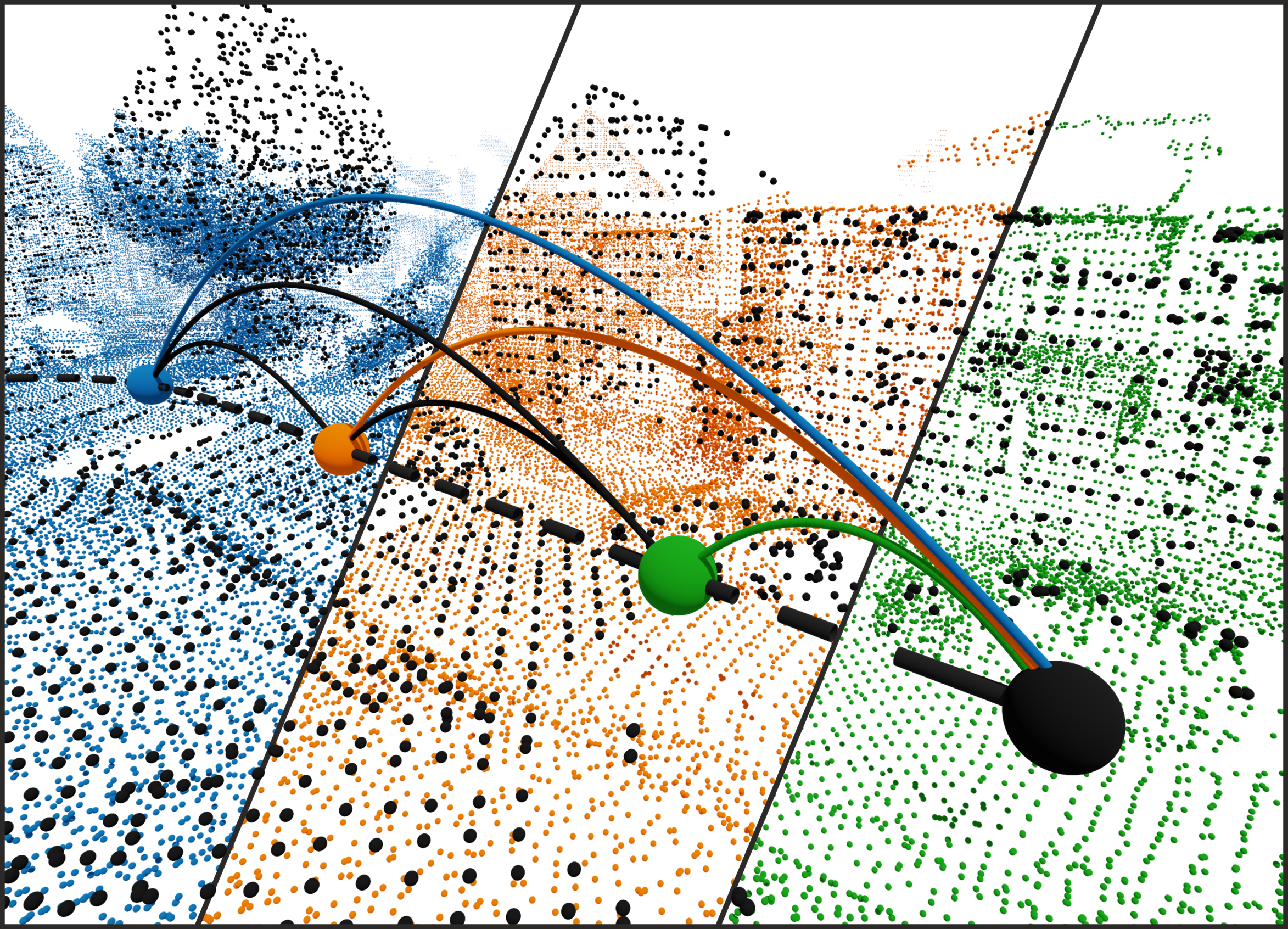}
    \caption{Schematic visualization of our approach, showing three submaps (blue, orange, green), the current aligned scan (black), as well as the underlying pose graph.}
    \label{fig:titleimage}
\end{figure}

Many modern methods utilize a scan-to-map registration approach, where the incoming lidar scan is registered against a static map built from prior scan alignments~\cite{vizzo2023ral,ferrari2024mad,lee2025genz}. This map is constructed incrementally as new scans are processed and serves as a fixed reference for all future registrations. Once inserted, the points of a scan are fixed and not updated. As a result, all registration errors introduced at an earlier stage remain in the map and affects all subsequent estimates. Fixed lag smoothing offers a means to mitigate this error accumulation by enabling the refinement of past poses using measurements acquired after their initial estimation~\cite{liang2022hierarchical,lv2023continuous}. Unlike online methods that compute the current pose using all available measurements up to the present time step, fixed lag smoothing targets the estimation of a pose that lies some time in the past. This allows for the incorporation of future measurements into the estimation of a past state, thereby improving both short-term accuracy and consistency. However, this delay poses a practical limitation for mobile agents that require timely and precise localization to respond rapidly to environmental changes.

\subsection{Contribution}
In this work we present a lidar-only odometry method that combines the retrospective refinement capabilities of the smoothing methodology with the online capabilities of conventional lidar odometry.
Instead of registering each scan against a single static map, our method registers the current scan against multiple overlapping submaps, leveraging redundant information to improve robustness and accuracy as visualized in Fig.~\ref{fig:titleimage}. This design enables accurate live pose estimates while simultaneously allowing for retroactive refinement of the map through global optimization. By updating the prior submaps' anchor points
with constraints obtained from new registrations, our approach reduces the error propagation inherent in fixed-map methods. We demonstrate that this framework achieves state-of-the-art performance on multiple benchmark datasets, consistently outperforming existing methods in both short-term as well as long-term accuracy, while maintaining real-time capability. 
In an ablation study, two key driving factors for the accuracy gain are identified: Firstly, the use of redundant information from our multitude-of-maps approach and secondly, the retrospective refinement of prior anchor points. 

\section{Related Work}

Point cloud registration and lidar odometry are fundamental problems in robotics and are essential for estimating motion from lidar data. These tasks have been studied extensively over the past decades. While the majority of existing approaches are based on the \ac{icp}~\cite{besl1992method} algorithm, alternative methods such as the \ac{ndt}~\cite{biber2003normal} have also been proposed. Registration strategies vary widely, ranging from point-based features~\cite{vizzo2023ral} and surface normals or other geometric features~\cite{wang2021floam, pan2021mulls,ferrari2024mad} to Gaussian mixtures~\cite{yokozuka2021litamin2, steuernagel2023point}, and, more recently, learned representations~\cite{dong2024lidar}. Modern state-of-the-art methods typically rely on a static map to infer motion, where the current lidar scan is registered against a map constructed from prior scans~\cite{vizzo2023ral,ferrari2024mad,lee2025genz} and therefore do not allow for retrospective changes of the map. 

The idea of leveraging future measurements for map and prior pose refinement is well established in the context of \ac{slam} and \ac{sam} approaches~\cite{shan2018legoloam,shan2020liosam}.
These methods, however, are often designed for post-processing, prioritizing accuracy and long-term consistency over real-time performance~\cite{grisetti2010hierarchical}. 
To this end, many approaches utilize a variety of sensor modalities in combination with lidar sensors to derive highly accurate estimates over prolonged durations~\cite{gentil2019in2lama,lv2023continuous,koide2024glim}. Lidar-only approaches on the other hand are less explored and often build upon augmenting odometry methods through the use of place-recognition or infrequent map-to-map refinements to increase accuracy~\cite{mendes2016icp,liang2022tight}. Such methods are often designed around the idea of fixed-lag smoothing~\cite{liang2022hierarchical} or pose post-processing~\cite{guadagnino2025kissslam} to derive highly accurate estimates after a fixed period of time.
Pure zero-lag lidar-only odometry methods capable of real-time pose estimation and retrospective map refinement remain uncommon, due to the increased computational complexity.

A recent method that addresses this gap is FORM~\cite{potokar2025form}.
In FORM, the authors propose the use of a pose graph optimization, where every single point-to-point and point-to-plane correspondence forms a constraint between the most current frame and the respective keyframe. Contrary to this, our approach utilizes a comparably sparsely connected graph, where only one constraint is derived per registration. 
The reliance on the densely connected graph maintained by FORM comes at the cost of increased computational load when constructing of the high-dimensional residual and Jacobian of the optimization problem, consuming the majority of available resources. Our approach, on the other hand, derives pose-pose constraints through robust \ac{icp} registrations which leads to a much smaller set of constraints and thus, a much faster construction of the optimization problem.
In fact, the time needed for constructing and solving the pose graph of our method only takes a small amount of time compared to the time needed for the derivation of the of the constraints via \ac{icp}.

\begin{figure*}
    \includegraphics[width=\textwidth]{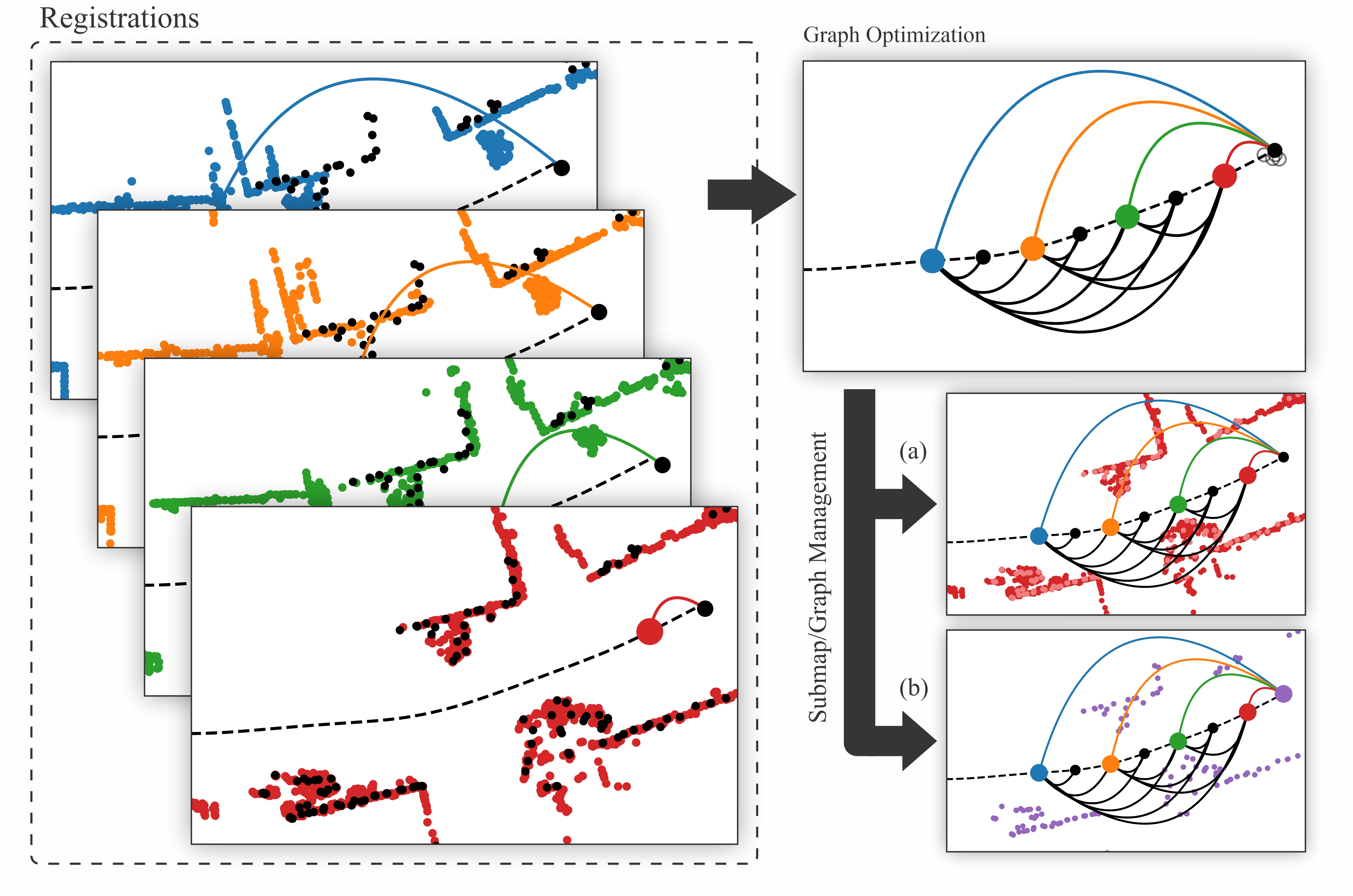}
    \caption{Depiction of our proposed method. Constraints from previous time steps are indicted in black. Left: The current scan (black) is registered independently against 4 submaps (colored dots), resulting in 4 constraints (colored edges). Top right: The four constraints are inserted into a pose graph and optimized, resulting in one pose estimate (black dot) for the current time step and a (slight) rearrangement of the other poses. Bottom right: Depending on the distance between the current pose estimate and the last submap, the points of the current scan are either (a): inserted into the last submap (red) or (b): create a new submap (purple).}
    \label{fig:pipeline}
\end{figure*}

\section{Method}\label{sec:method}
Given a sequence of ego-centered lidar scans $\{\set P_{1},...,\set P_{t}\}$ from a moving platform up to the current time $t$, the goal is to estimate the 6 \ac{dof} pose $\state_{t} \in \SE 3$ at the current time $t$.
At its core, our method relies on the extensively studied \ac{icp} algorithm for the alignment of two point clouds. However, in contrast to other modern lidar odometry methods, our approach utilizes not a single, but multiple registrations against a set of overlapping maps.  This idea is depicted in Fig.~\ref{fig:pipeline}, where the most current lidar scan (black) is registered against four (overlapping) maps (blue, orange, green, red). Every registration forms one constraint (colored edge) from the map's anchor point to the most current lidar scan. Adding the newly derived edges into a maintained pose graph enables us to not only find an optimal pose estimate for the most current scan, but also (slightly) rearranges prior poses, increasing the local consistency of the trajectory and the accuracy of future registrations. The result is a lidar-only odometry approach that enables online and real-time lidar odometry with the ability to (retrospectively) refine the map. 

\subsection{Notation}
For the remainder of this paper we will adopt the following notation:
States and measurements $\state_{t}, \meas_{t} \in \SE 3$ will be denoted by bold, lower case letters. Lidar scans $\set P_{t}$, i.e., point clouds, will be denoted with calligraphic, upper case letters. In both cases, a subscript $(\cdot)_t$ indicates the time step. A superscript $(\cdot)^k$ of a state, measurement or point cloud indicates its reference frame. States and measurements without superscript are expressed in the global reference frame, i.e., $\state_t=\state_t^0$, whereas point clouds without superscripts are expressed in the ego-frame to the vehicle, i.e., $\set P_t=\set P_t^t$.

\subsection{Scan Preprocessing}
For scan preprocessing, we utilize a double voxel hash map approach as proposed by~\cite{vizzo2023ral}. The raw scan of the current time step $t$ is first deskewed (if necessary) and inserted into a 3D voxel hash map with a voxel size of $v_{\text{map}}=\SI{0.4}{\meter}$, keeping only the first inserted point per voxel. The result is a subset $\set H_{t} \subseteq \set P_{t}$ of the raw scan $\set P_{t}$. $\set H_{t}$ is then subsampled a second time with an increased voxel size $v_{\text{ICP}}=\SI{1.8}{\meter}$, yielding $\set I_{t} \subseteq \set H_{t} \subseteq \set P_{t}$. The stronger subsampled scan $\set I_{t}$ will be used for the registration and the denser $\set H_{t}$ will be used to update or create new submaps.

\subsection{Motion Prediction}
Similar to other lidar-odometry methods, we utilize a constant motion model for motion prediction. This model is both used for scan deskewing and for the initialization of the \ac{icp} registrations. Given the last two poses $\state_{t-1}, \state_{t-2} \in \SE 3$ in a global reference frame, we can calculate the motion $\dot{\state}_{t-1} \in \SE 3$ of our platform at the last time step with

\begin{equation}
    \dot{\state}_{t-1} = \state_{t-2}^{-1}\oplus\state_{t-1}
    \enspace,
    \label{eq:motion_velocity}
\end{equation}
with $\state^{-1}$ denoting the inverse of $\state$ and $\oplus$ denoting the pose composition operator as described in~\cite{blanco2021tutorial}. Using the assumption of constant motion and constant time between frames, we can calculate the predicted pose at the current time $t$ with

\begin{equation}
    \hat{\state}_{t} = \state_{t-1}\oplus \dot{\state}_{t-1}
    \enspace.
    \label{eq:motion_predicted_pose}
\end{equation}

\subsection{Iterative Closest Point}\label{sec:icp}
Registrations between point clouds are computed using a robust \ac{icp} algorithm.
Its application to two point clouds $\set A$ and $\set B$ is denoted as
\begin{equation}
    \Delta \measurement, \information = \text{ICP}(\set A, \set B)
    \enspace.
    \label{eq:icp}
\end{equation}
The result $\Delta \measurement \in \SE 3$ is the transformation that aligns $\set A$ to $\set B$ and $\information$ will be used as the information matrix for the pose graph optimization in Section~\ref{sec:smoothing}. 
We use the Hessian method~\cite{bengtsson2003robot,censi2007accurate} as an approximation for the (inverse) covariance of the registration.

The general implementation of \ac{icp} can be summarized as the repeated calculation of a set of correspondences ${\set C \subseteq \set A \times \set B}$, followed by the minimization of the distances of these correspondences. These two steps are repeated until a local minimum is reached.
For the calculation of correspondences we use a standard \ac{nn} search for all points $\mathbf{a} \in \set A$ to any point $\mathbf{b} \in \set B$ in a maximum search radius of $d_{max}=3\si{\meter}$. 

For the optimization of the correspondences we use the Gauss-Newton method to minimize

\begin{equation}
    \Delta\measurement_{i} = \underset{\T \in \SE 3}{\text{argmin}} \sum_{(\vec a,\vec b) \in \set C_{i}} \rho(||\T\vec a - \vec b||_2)
    \enspace,
    \label{eq:icp_optimization}
\end{equation}
with the subscript $(\cdot)_{i}$ indicating the iteration of the \ac{icp} algorithm.
Instead of minimizing the squared distance, a robust kernel
\begin{equation}
    \rho(e) = \frac{e^2/2}{\tau+e^2}
    \enspace,
    \label{eq:icp_kernel}
\end{equation}
as proposed in~\cite{chebrolu2021adaptive}, is used to reduce the effect of outliers on the registration. Here, the parameter $\tau$ controls the distance beyond which correspondences are considered as outliers. It is adaptively chosen based on the magnitude of prior \ac{icp} corrections similarly to~\cite{vizzo2023ral}. 

Correspondence calculation and optimization are repeated until the convergence of $\Delta\measurement_{i}$, i.e., $||\Delta\measurement_{i}||_2 < 10^{-4}$ or until a maximum of $100$ iterations is reached. Between iterations the source point cloud $\set A$ is transformed with the last increment $\Delta\measurement_{i}$.
The final result of the \ac{icp} algorithm is then the product of all increments computed as

\begin{equation}
    \Delta\measurement = \Delta\measurement_{N} \oplus \Delta\measurement_{N-1} \oplus ... \oplus \Delta\measurement_{1}
    \enspace.
\end{equation}

\subsection{Multiple Scan-to-Scan Registration}\label{sec:micp}
An integral part of the proposed algorithm is to maintain not only a single local map, but instead a set of aligned submaps, against which multiple registrations can be carried out.
Our method maintains a set of submaps $\mathfrak{K} = \{\mathcal{K}_{\alpha_1},...,\mathcal{K}_{\alpha_N}\}$ 
with their respective anchor poses $\{\state_{\alpha_1},...,\state_{\alpha_N}\}$. We register the current preprocessed lidar scan $\set I_{t}$ independently to all submaps $\mathcal{K}_{\alpha_i} \in \mathfrak{K}$.
For this, we first express the predicted pose $\hat{\state}_{t}$ from our motion model \eqref{eq:motion_predicted_pose} 
in the reference frame of the submap under consideration $\mathcal{K}_{\alpha_i}$ with
\begin{equation}
    \hat{\state}_{t}^{\alpha_i} = \state_{\alpha_i}^{-1}\oplus\hat{\state}_{t}
    \enspace.
    \label{eq:rel}
\end{equation}
The result $\hat{\state}_{t}^{\alpha_i}$ is used as an initial guess for a registration between the current scan $\set I_{t}$ and the submap $\mathcal{K}_{\alpha_i}$

\begin{equation}
    \Delta \measurement_{t}^{\alpha_i}, \information_{t}^{\alpha_i} = \text{ICP}(\hat{\state}_{t}^{\alpha_i} \cdot \set I_{t}, \mathcal{K}_{\alpha_i})
    \enspace,
\end{equation}
resulting in a correction $\Delta \measurement_{t}^{\alpha_i}$ of the initial guess $\state_{t}^{\alpha_i}$, as well as the estimated information matrix $\information_{t}^{\alpha_i}$. 
The product operator $(\cdot) : \SE 3 \times \mathbb{R}^{N\times3}\rightarrow \mathbb{R}^{N\times3}$ indicates the element-wise application of a transformation to all points of a point cloud.
The corrected (relative) pose $\measurement_{t}^{\alpha_i}$ of this registration can then be obtained with 
\begin{equation}
    \measurement_{t}^{\alpha_i}=\Delta \measurement_{t}^{\alpha_i} \oplus \hat{\state}_{t}^{\alpha_i}
    \enspace.
\end{equation}
This procedure is applied for all submaps, resulting in a set of pose-information tuples $\mathcal{C}_{t} = \{(\measurement_{t}^{\alpha_1},\information_{t}^{\alpha_1}),...,(\measurement_{t}^{\alpha_N},\information_{t}^{\alpha_N})\}$ which will be added to the pose graph optimization as new constraints, connecting the pose of the current time step to the rest of the graph as depicted in Fig.\ref{fig:pipeline}.
\subsection{Pose Graph Optimization}\label{sec:smoothing}
The goals of the pose graph optimization are firstly, to derive a singular (optimal) pose estimate for the current time step $t$ and secondly, to refine the configuration of prior poses to increase the accuracy of future registrations. For this, we maintain a small, densely connected pose graph throughout the execution of our method. This includes a set of constraints (black edges in Fig.~\ref{fig:pipeline}) and poses (colored and black dots) up to the last time step $t-1$. Formally we denote them with $\mathcal{C}_{\alpha_i:t-1}$ and $\mathbf{X}_{\alpha_i:t-1}$ respectively. The current scan is registered against the set of submaps (colored dots) as described in Sec~\ref{sec:micp}. The resulting constraints $\mathcal{C}_{t}$ (colored edges) and an initial guess for the current pose $\state_{t} = \state_{\alpha_N} \oplus \measurement_{t}^{\alpha_N}$ are inserted into the graph, resulting in $\mathcal{C}_{\alpha_i:t} = \mathcal{C}_{\alpha_i:t-1} \cup \mathcal{C}_{t}$ and $\mathbf{X}_{\alpha_i:t} = \mathbf{X}_{\alpha_i:t-1} \cup \state_{t}$.

The (new) best configuration of poses $\mathbf{X}^*_{\alpha_i:t}$ can then be obtained by solving the minimization problem

\begin{equation}
    \mathbf{X}^*_{\alpha_i:t} = \underset{\mathbf{X}}{\text{argmin}} \sum_{(\measurement_{ij},\information_{ij}) \in \mathcal{C}_{\alpha_i:t}} \theta(\mathbf{e}_{ij}) \cdot \mathbf{e}_{ij}^T \information_{ij} \mathbf{e}_{ij}
    \enspace ,
    \label{eq:pose_graph}
\end{equation}
where we redefine

\begin{align}
    \information_{ij} &\overset{\text{\tiny{def}}}{=} \information_{i}^{j} \enspace, \\
    \measurement_{ij} &\overset{\text{\tiny{def}}}{=} \measurement_{i}^{j} \enspace, \\
    \mathbf{e}_{ij} &\overset{\text{\tiny{def}}}{=} (\measurement_{i}^{j})^{-1} \oplus (\state_{j}^{-1} \oplus \state_{i}) \enspace,
    \label{eq:shorthands}
\end{align}
for the information matrices, measurements and the optimized error function in order to conform with standard pose graph notation~\cite{grisetti2010hierarchical,grisetti2010tutorial}. The essential difference to standard pose graph optimization is the introduction of a gating mechanism 

\begin{equation}
    \theta(\mathbf{e}) = 
    \begin{cases}
        1 & ||\text{log}(\mathbf{e})^\vee||_2 \leq g_{max} \\
        0 & \, \text{otherwise}
        \enspace ,
        \end{cases}
    \label{eq:theta_e}
\end{equation}
that excludes erroneous constraints above a hard threshold of $g_{max}$ from contributing to the optimization. 
The $\text{log}(\cdot)^\vee: \SE3 \rightarrow \mathbb{R}^6$ operator is the logarithm that maps a transformation into its twist vector as defined in~\cite{blanco2021tutorial}.
The primary motivation is to enhance robustness against erroneous registrations arising from low-overlap scan alignments. In particular, when the current scan is registered to distant submaps, the resulting overlap may be minimal. While such alignments are typically unproblematic in geometrically rich environments, they may result in erroneous registrations in regions with insufficient structural detail, such as tunnels. We find a value of $g_{max}=v_{map}$, i.e., the voxel size of the map, to yield the best results and used it throughout the experiments. Because only exactly one node and few constraints are inserted into to pose graph at any time step, the optimization of~\eqref{eq:pose_graph} converge fast, typically terminating in under $5$ iterations of the Gauss-Newton method. 
Finally, we update the system state by updating the current pose estimates $\mathbf{X}_{\alpha_i:t} \leftarrow \mathbf{X}^*_{\alpha_i:t}$ with the optimized configuration and, henceforth, all future references to $\state_t$ denote references to the optimized poses instead.

\subsection{Submap and Graph Management}\label{sec:submap_management}
The computational complexity of our method is dominated by the multiple \ac{icp} registrations step in Sec.~\ref{sec:micp}. Since this cost scales linearly with the number of submaps, limiting the number of independent registrations is critical for real-time performance. 
Instead of creating one submap per lidar scan, we assume negligible error over very short distances ($\SI{5}{\meter}-\SI{10}{\meter}$, depending on the maximum lidar range). Let $d = {||\text{log}(\state_t^{-1} \oplus \state_{\alpha_N})^\vee||_2}$ denote the distance between the optimized current pose and optimized anchor pose of the last submap (red dot in Fig.~\ref{fig:pipeline}).
If $d < k_{min}$, the subsampled point cloud $\set H_t$ is expressed in the local coordinate frame of the last submap with $\set H_t^{\alpha_N} = \state_{\alpha_N}^{-1} \oplus \state_t \cdot \set H_t$ and then merged with the last submap $\set K_{\alpha_N} \leftarrow \set K_{\alpha_N} \cup \set H_t^{\alpha_N}$. 
In this case, $\state_t$ is not considered as a new submap. Conversely, if $d \geq k_{min}$, $\set H_t$ establishes a new submap $\set K_t = \set H_t$ (purple dot) and is added to $\mathfrak{K}$; future registrations will be carried out against it.

To keep the computational complexity of the approach bounded over time, we remove old submaps and poses $\state_i$ from the graph if their distance to the current pose $\state_t$ becomes too large.
We prune all nodes $\state_i \in \mathbf{X}_{\alpha_i:t}$ (and their accompanying edges) that satisfy ${||\text{log}((\state_t)^{-1} \oplus \state_i)^\vee||_2} > k_{max}$. In all our experiments we set $k_{max}=l_{max}/2$ and $k_{min}=k_{max}/5$ according to the maximum distance of the lidar $l_{max}$, which means that our approach maintains at least five submaps when driving straight and more submaps when moving in tight circles. While the maximum number of submaps is theoretically unbounded, this was not a problem in the evaluated datasets. Depending on the domain, it may be advisable to limit the number of submaps to a suitable amount.

\section{Evaluation}\label{sec:evaluation}
We have conducted experiments on three different real-world lidar odometry datasets to judge the performance of our method. These are the KITTI~\cite{geiger2012we}, 
MulRan~\cite{kim2020mulran} and the Odyssey~\cite{kurda2025odyssey} datasets.
The metrics used in our evaluation are the popular $\text{RPE}_{100}$ and KITTI-metric\footnote{https://www.cvlibs.net/datasets/kitti/eval\_odometry.php}.
The $\text{RPE}_{100}$ measures the amount of error accumulated over distances of $100 \si{\meter}$ and can therefore be interpreted as a measure of the short-term drifting behavior. The KITTI-metric on the other hand measures the average error over a set of longer distances $[100,200,...,800]$. Consequently, it can be regarded as a measure for medium- to long-term error behavior. All experiments were conducted on a system running Ubuntu 22.04 with an Intel Core i7-13700K.

\begin{table}
\caption{Ablation study on the MulRan~\cite{kim2020mulran} dataset. Best results are marked in \textbf{bold}.}
\label{tab:4}
\centering
\resizebox{\columnwidth}{!}{%
\begin{tabular}{l|rrrr|rrrr}
\specialrule{.17em}{.3em}{.3em} 
 & \multicolumn{4}{c|}{KITTI} & \multicolumn{4}{c}{$\text{RPE}_{100}$} \\
 \specialrule{.1em}{.3em}{.3em}  
\multirow{2}{*}{\centering} & 
\multicolumn{1}{c}{Ours} & 
\multicolumn{1}{c}{GO} & 
\multicolumn{1}{c}{\raisebox{-.5\height}{SI}} & 
\multicolumn{1}{c|}{GO} & 
\multicolumn{1}{c}{Ours} & 
\multicolumn{1}{c}{GO} & 
\multicolumn{1}{c}{\raisebox{-.5\height}{SI}} & 
\multicolumn{1}{c}{GO} \\

 &
 
 \multicolumn{1}{c}{(Full)} & 
 \multicolumn{1}{c}{LA} &  
 & 
 \multicolumn{1}{c|}{NO} &  
\multicolumn{1}{c}{(Full)} & 
 \multicolumn{1}{c}{LA} & 
 & 
 \multicolumn{1}{c}{NO}\\ 
 \specialrule{.1em}{.3em}{.3em} 
DCC01 & \textbf{2.56} & 2.61 & 2.87 & 3.88 & \textbf{3.47} & 3.50 & 3.67 & 4.04 \\
DCC02 & \textbf{2.06} & 2.09 & 2.69 & 4.08 & \textbf{2.48} & 2.48 & 2.77 & 3.21 \\
DCC03 & \textbf{1.67} & 1.78 & 2.23 & 3.53 & \textbf{2.46} & 2.54 & 2.83 & 3.18 \\
KAIST01 & \textbf{1.97} & 1.98 & 2.25 & 3.16 & \textbf{2.31} & \textbf{2.31} & 2.42 & 2.70 \\
KAIST02 & \textbf{1.91} & 1.92 & 2.23 & 3.55 & 2.29 & \textbf{2.28} & 2.42 & 2.83 \\
KAIST03 & 2.28 & \textbf{2.27} & 2.54 & 3.87 & 2.57 & \textbf{2.56} & 2.68 & 3.00 \\
Riverside01 & \textbf{3.14} & 3.19 & 3.38 & 4.21 & \textbf{2.77} & 2.78 & 2.83 & 2.92 \\
Riverside02 & 3.06 & \textbf{2.98} & 3.29 & 4.59 & 2.73 & \textbf{2.59} & 2.68 & 3.09 \\
Riverside03 & \textbf{1.96} & 2.00 & 2.26 & 3.16 & \textbf{2.09} & \textbf{2.09} & 2.14 & 2.40 \\
Sejong01 & \textbf{3.53} & 4.28 & 4.58 & 7.78 & \textbf{2.66} & 3.06 & 2.87 & 3.49 \\
Sejong02 & \textbf{3.97} & 4.10 & 5.08 & 8.61 & \textbf{2.96} & 2.97 & 3.14 & 3.72 \\
Sejong03 & \textbf{4.30} & 4.48 & 5.22 & 8.72 & \textbf{3.20} & 3.32 & 3.33 & 4.17 \\
\specialrule{.1em}{.3em}{.3em} 
mean & \textbf{2.70} & 2.81 & 3.22 & 4.93 & 2.67 & 2.71 & 2.81 & 3.23 \\
\specialrule{.17em}{.3em}{.3em} 
\end{tabular}
}
\end{table}
\begin{table}
\caption{Results on the KITTI-odometry~\cite{geiger2012we} dataset. Best results are marked in \textbf{bold}.}
\label{tab:res_kitti}
\centering
\resizebox{\columnwidth}{!}{%
\begin{tabular}{l|rrrrr|rrrrr}
\specialrule{.17em}{.3em}{.3em} 
 & \multicolumn{5}{c|}{KITTI} & \multicolumn{5}{c}{$\text{RPE}_{100}$} \\
 \specialrule{.1em}{.3em}{.3em}  
\multirow{2}{*}{\centering} & \multicolumn{1}{c}{\raisebox{-.5\height}{Ours}} & \multicolumn{1}{c}{KISS-} & \multicolumn{1}{c}{MAD-} & \multicolumn{1}{c}{GenZ-} & \multicolumn{1}{c|}{KISS-} & \multicolumn{1}{c}{\raisebox{-.5\height}{Ours}} & \multicolumn{1}{c}{KISS-} & \multicolumn{1}{c}{MAD-}  & \multicolumn{1}{c}{GenZ-} & \multicolumn{1}{c}{KISS-}\\
&    & \multicolumn{1}{c}{ICP} & \multicolumn{1}{c}{ICP} & \multicolumn{1}{c}{ICP} & \multicolumn{1}{c|}{SLAM} &    & \multicolumn{1}{c}{ICP} & \multicolumn{1}{c}{ICP} & \multicolumn{1}{c}{ICP} & \multicolumn{1}{c}{SLAM}\\
\specialrule{.1em}{.3em}{.3em}   
00 & 0.55 & 0.52 & 0.76 & \textbf{0.51} & 0.52 & \textbf{0.80} & \textbf{0.80} & 0.86 & \textbf{0.80} & \textbf{0.80} \\
01 & \textbf{0.63} & 0.78 & 0.99 & 0.92 & 0.79 & \textbf{0.79} & 0.83 & 0.84 & 0.88 & 0.84 \\
02 & 0.57 & 0.53 & 0.80 & \textbf{0.51} & 0.54 & \textbf{0.78} & 0.79 & 0.79 & 0.80 & 0.79 \\
03 & \textbf{0.68} & \textbf{0.68} & 2.19 & 0.72 & \textbf{0.68} & \textbf{0.83} & 0.85 & 2.18 & 0.88 & 0.85 \\
04 & \textbf{0.35} & 0.39 & 0.63 & 0.39 & 0.38 & \textbf{0.42} & \textbf{0.42} & 0.52 & 0.43 & 0.43 \\
05 & \textbf{0.26} & 0.34 & 0.51 & 0.27 & 0.34 & \textbf{0.43} & 0.48 & 0.45 & 0.45 & 0.49 \\
06 & \textbf{0.26} & 0.28 & 0.43 & 0.28 & 0.31 & 0.43 & 0.48 & \textbf{0.39} & 0.48 & 0.50 \\
07 & 0.33 & 0.37 & 0.44 & \textbf{0.31} & 0.37 & \textbf{0.48} & 0.55 & 0.54 & \textbf{0.48} & 0.54 \\
08 & \textbf{0.80} & 0.81 & 1.22 & 0.83 & 0.83 & \textbf{1.33} & 1.35 & 1.41 & 1.35 & 1.35 \\
09 & \textbf{0.49} & 0.53 & 1.17 & 0.51 & 0.55 & \textbf{0.59} & 0.61 & 0.74 & 0.63 & 0.61 \\
10 & 0.62 & 0.53 & 1.24 & \textbf{0.46} & 0.50 & 0.72 & \textbf{0.70} & 0.89 & 0.68 & 0.69 \\
\specialrule{.1em}{.3em}{.3em} 
mean & \textbf{0.50} & 0.52 & 0.94 & 0.52 & 0.53 & \textbf{0.69} & 0.71 & 0.87 & 0.71 & 0.72 \\
\cmidrule{1-6}
FPS & 50.55 & \textbf{123.09} & 19.90 & 68.45 & 91.00 \\
\specialrule{.17em}{.3em}{.3em} 
\end{tabular}
}
\end{table}

\subsection{Ablation Studies} \label{sec:ablation}
An ablation study was conducted to identify the sources of accuracy gain in the proposed method and assess the impact of key design choices. The method was modified in three ways to evaluate the contribution of each component.
Each variation only affected a single component, leaving the rest of the algorithm as previously described. The results are shown in Table~\ref{tab:4}.
First, GO-LA allows only the current pose to be optimized in the pose graph optimization and thus disables retrospective refinements of the map. Second, GO-NO disables the graph optimization entirely. Third, SI forces the system to maintain only a single submap at the origin, effectively reverting to a scan-to-map approach similar to KISS-ICP. To keep the size of the map in this case bounded, we removed points from the map that were too far away form the current pose estimate.

GO-NO performs the worst among all variants, followed by SI, both of which operate as scan-to-map odometry pipelines similar to KISS-ICP. The poor performance of GO-NO arises because the current pose is determined exclusively from the registration against the most recent submap, meaning inserting a new submap effectively resets the map and therefore disrupts short-term accuracy. In contrast, SI maintains a single submap at the origin that is never reset, therefore providing more stable registrations than GO-NO. Compared to the complete pipeline, it lacks both the retrospective refinement of previous submaps as well as the redundant information from the multiple alignments, resulting in an overall worse accuracy.
GO-LA demonstrates a significant performance improvement over SI, reducing the KITTI error by $13\%$ and the $\text{RPE}_{100}$ by $4\%$. The full pipeline further improves upon GO-LA by $4\%$ on the KITTI metric and $1\%$ on the $\text{RPE}_{100}$. These improvements reveal that the multi-submap registration approach and the retrospective refinement provide significant benefits over longer durations. Conversely, their impact is more negligible over short distances, as indicated by the smaller relative gains on the $\text{RPE}_{100}$ metric.

\subsection{Results and Comparison}

We compare the results of our method against four publicly available state-of-the-art lidar odometry and lidar \ac{slam} methods: KISS-ICP~\cite{vizzo2023ral}, MAD-ICP~\cite{ferrari2024mad}, GenZ-ICP~\cite{lee2025genz} and KISS-SLAM~\cite{guadagnino2025kissslam}. While KISS-ICP is based on pure point-to-point registration, MAD-ICP additionally utilizes the surface-normal information of the local neighborhood to improve accuracy in structured environments. In contrast, GenZ-ICP dynamically combines both point-to-point and point-to-plane registration within a single optimization step, adaptively selecting the appropriate metric based on the local geometry. KISS-SLAM is a full~\ac{slam} system built on top of KISS-ICP that additionally uses place-recognition for loop closure detection. Contrary to all other methods, the pose estimates produced by KISS-SLAM are smoothed through a final pose graph optimization at the end of each sequence, which means that future information from time $t+n$ is used to derive poses at time $t$. The results of our evaluation on the KITTI, MulRan and Odyssey datasets are summarized in Tables~\ref{tab:res_kitti},\ref{tab:res_mulran}, and~\ref{tab:res_odyssey}. 

In terms of processing speed, KISS-ICP emerges as the fastest method, followed by KISS-SLAM, GenZ-ICP, our proposed method, and finally MAD-ICP. While slower than the top contenders, our method achieves a significant accuracy advantage, especially on the newer MulRan and Odyssey datasets. On the KITTI dataset, all methods excluding MAD-ICP performed comparably. The only exception is sequence 01, where our method achieved an $18\%$ reduction in the KITTI metric over KISS-ICP. Overall, our method marginally outperformed the second-best method (KISS-ICP) by $4\%$ on the KITTI metric and $3\%$ on the $\text{RPE}_{100}$ on this dataset. On the MulRan and Odyssey datasets, our method consistently outperformed the competition, reducing overall errors by $10\%$ and $23\%$ in KITTI metric as well as $2\%$ and $11\%$ in $\text{RPE}_{100}$.

\begin{figure}
    \centering
    \includegraphics[width=0.49\textwidth]{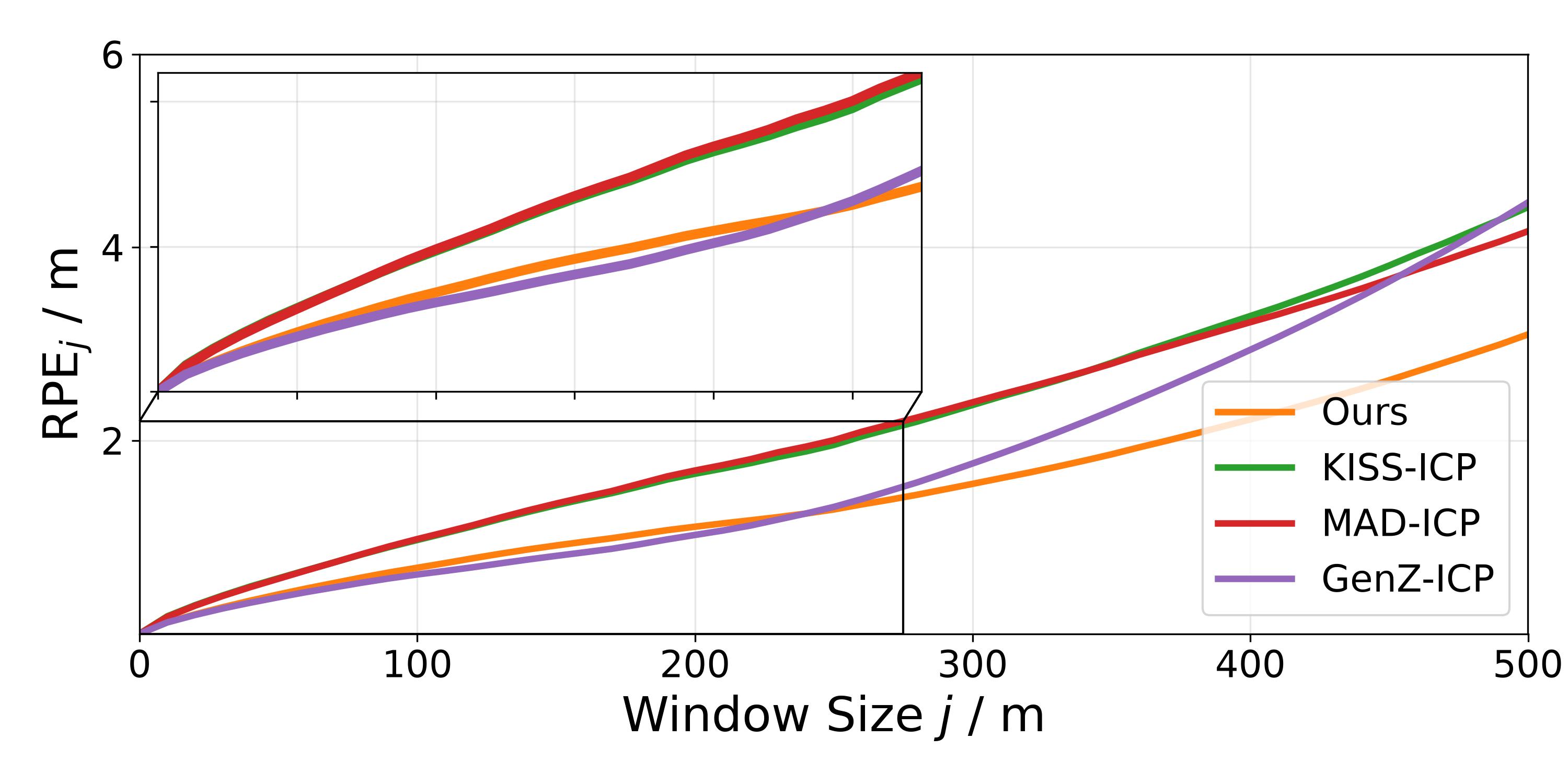}
    \caption{$\text{RPE}_{j}$ over a range of window sizes $j$ on the Odyssey dataset. The \texttt{Highway} trajectory was omitted in the calculation to ensure a fair comparison between GenZ-ICP and the other methods.}
    \label{fig:4}
\end{figure}

A closer sequence-by-sequence inspection of the results on the Odyssey dataset reveals a comparably bad accuracy of GenZ-ICP on the \texttt{Highway} sequences. We found the method to consistently fail to correctly estimate the orientation during the traversal of a bridge at high speeds. The failure of GenZ-ICP on all three repetitions points towards a systematic problem of the method in this situation specifically. Removing the \texttt{Highway} sequence from the calculation of the mean results in an error of $0.76$ for the KITTI metric and $0.6$ for the $\text{RPE}_{100}$, surpassing our method in terms of $\text{RPE}_{100}$ by $11\%$ but still falling $7\%$ short in terms of the KITTI metric. This shows that, while capable of accurate motion estimation over short periods, the accuracy of GenZ-ICP falls off over longer distances. Fig.~\ref{fig:4} visualizes this fact by evaluating the $\text{RPE}_{j}$ over a variety of window sizes $j$. From this it is observable that GenZ-ICP indeed is the best performing method for window sizes $\leq 250$, but drops off faster over increasing distances than the other methods.

\begin{figure*}
    \centering
    \subfloat[\centering \texttt{00}]{\includegraphics[width=0.33\textwidth]{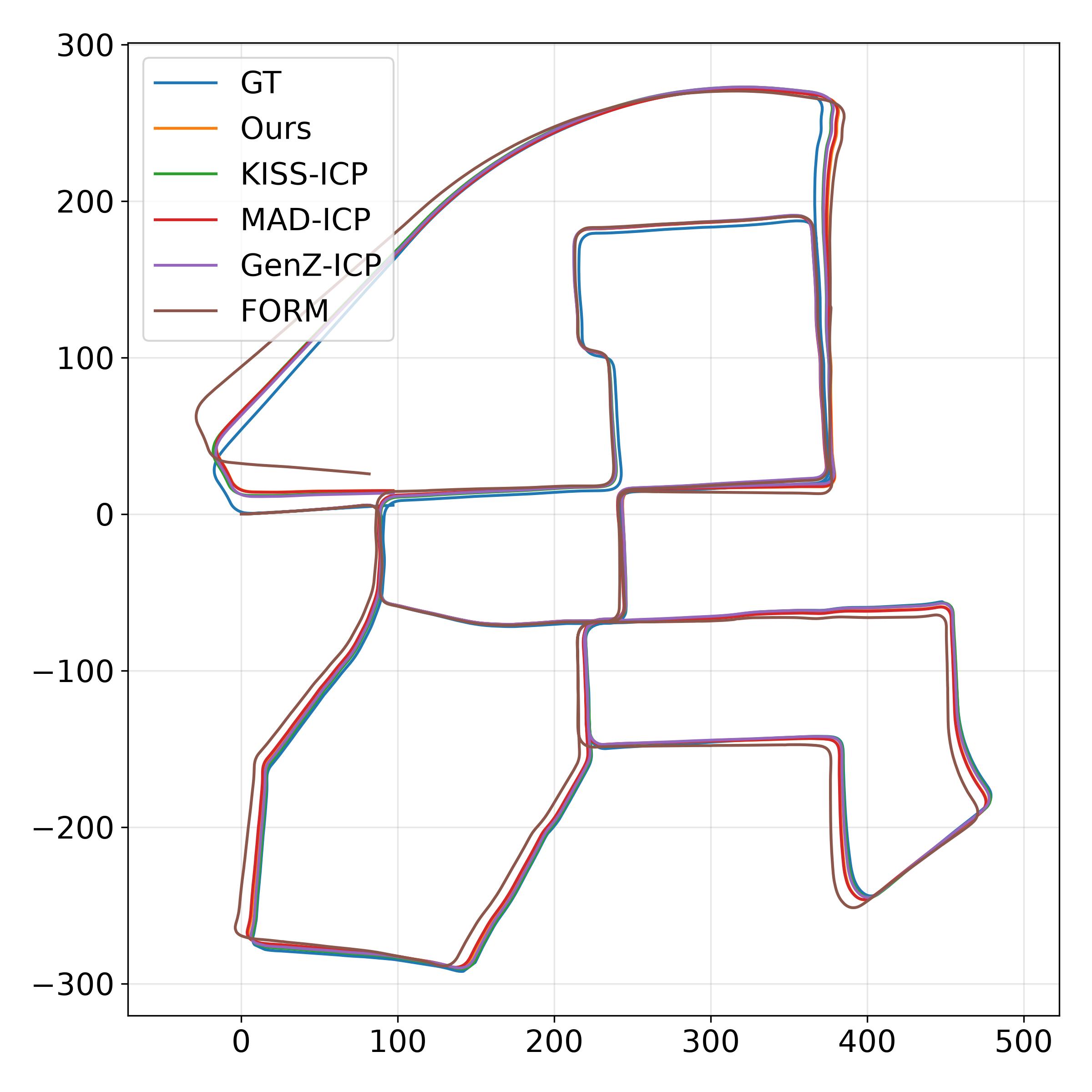}\label{fig:3:kitti_00}}
    \subfloat[\centering \texttt{ForestRoad1}]{\includegraphics[width=0.33\textwidth]{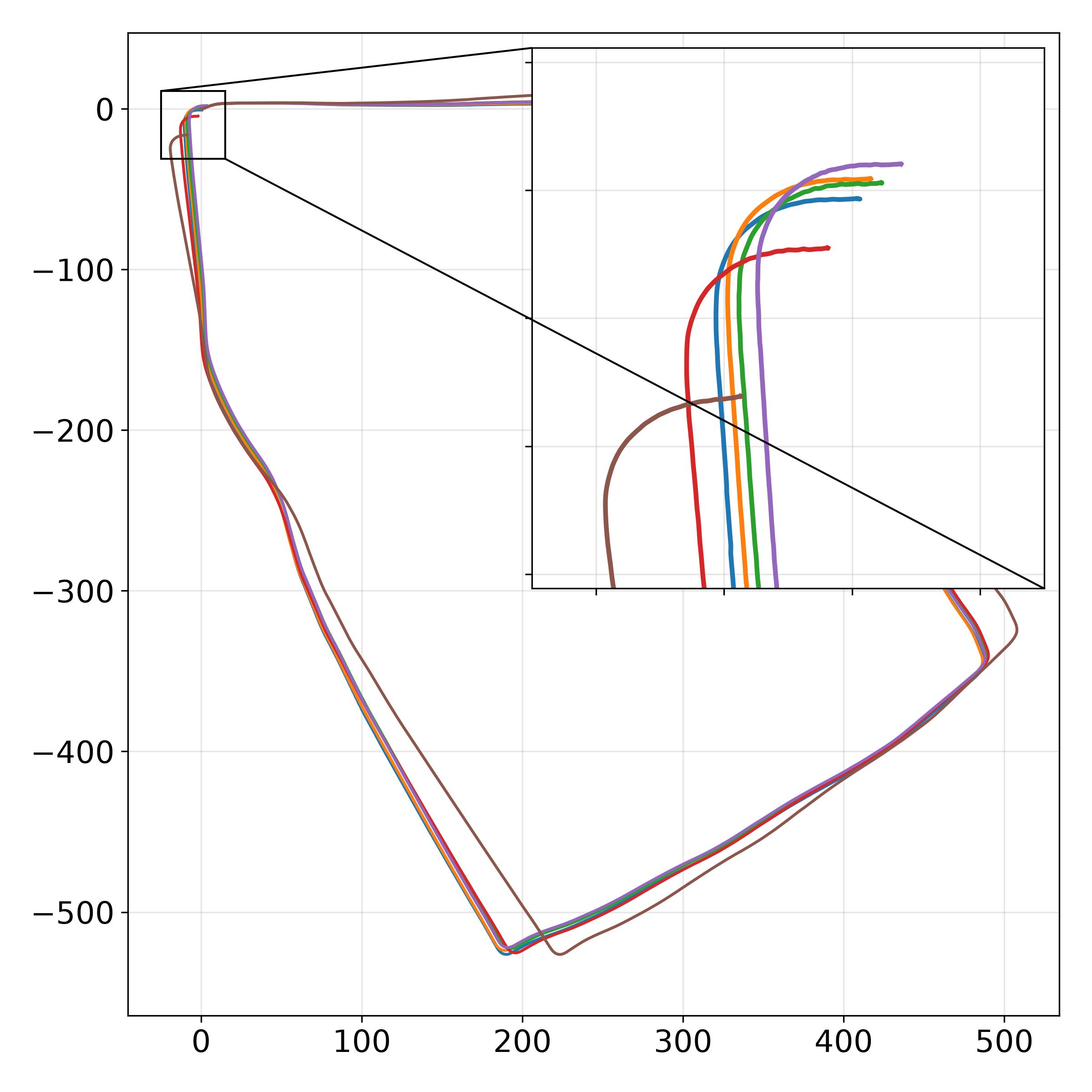}\label{fig:3:odyssey_forestroad1}}
    \subfloat[\centering \texttt{ParkingGarage2}]{\includegraphics[width=0.33\textwidth]{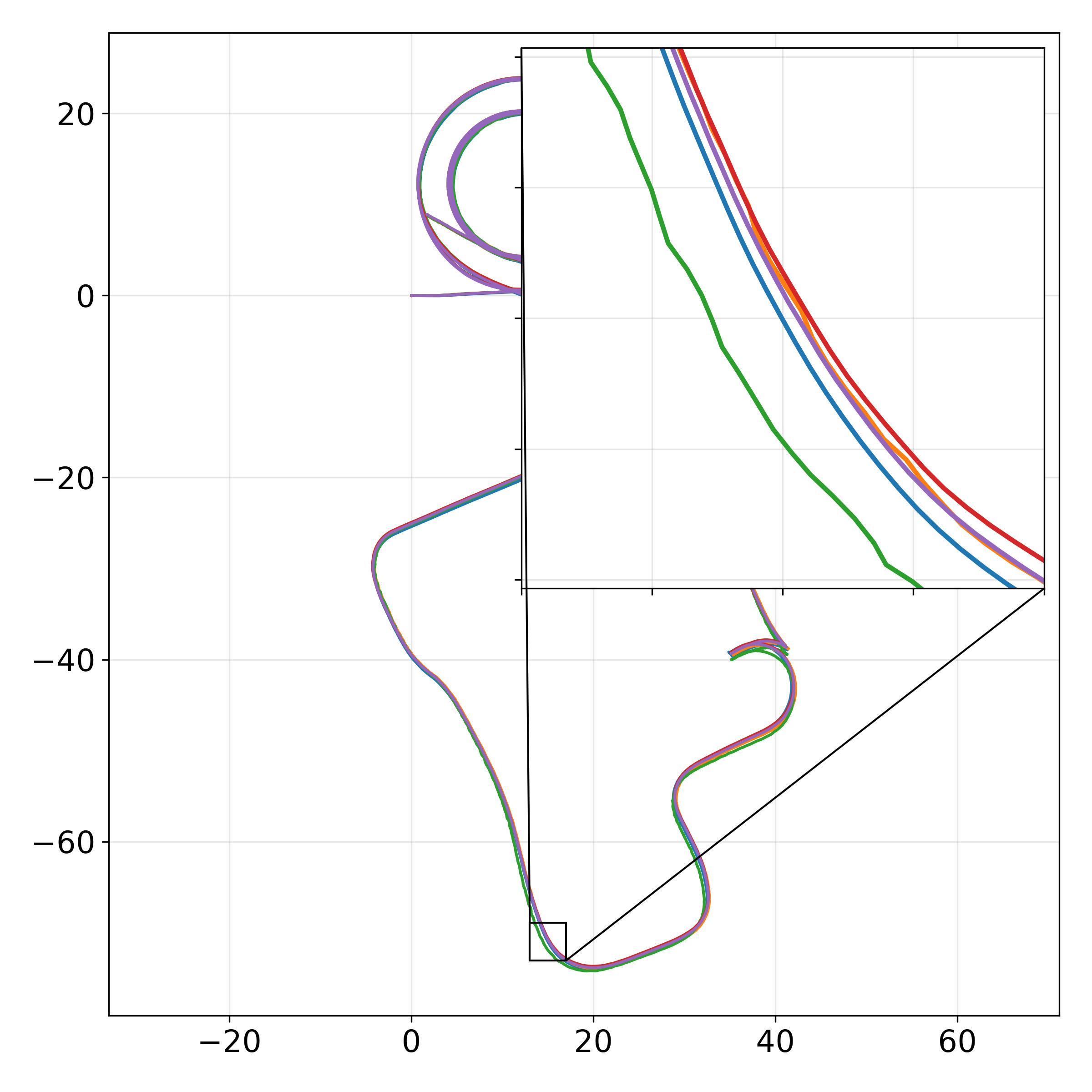}\label{fig:3:odyssey_parkinggarege02}}
    \caption{Qualitative results of several lidar odometry methods on the KITTI and Odyssey datasets. Similar to the point-to-plane methods MAD-ICP and GenZ-ICP, our method provides smooth motion estimates.}
    \label{fig:3}
\end{figure*}

To further assess the quality of the estimated trajectories, a visual/qualitative analysis of the sequences across all datasets was conducted. On the KITTI dataset, no discernible difference in smoothness or drift patterns was observable (Fig.~\ref{fig:3:kitti_00}). On the MulRan and Odyssey dataset on the other hand, we found KISS-ICP to often produce jagged trajectories, especially in slow-moving situations such as \texttt{ParkingGarage} (Fig.~\ref{fig:3:odyssey_parkinggarege02}), while the two point-to-plane based methods, MAD-ICP and GenZ-ICP, as well as our method, produced much smoother estimates. For our method, we traced the cause for the increased smoothness to the pose-graph optimization of the multiple independent registrations. When executing our method in the SI configuration (as described in Sec.~\ref{sec:ablation}), we observed a similar jaggedness to KISS-ICP. This is expected behavior as the SI configuration essentially reverts our method back into to conventional scan-to-map lidar odometry method similar to KISS-ICP. On the other hand, when executing our method with pose graph optimization enabled using the GO-LA configuration, this jaggedness disappeared, showcasing the smoothing effect of our multiple submap registration approach.
Overall, our method achieves superior robustness and accuracy in the automotive domain compared to state-of-the-art alternatives, at the cost of a moderate increase in computational complexity over KISS-ICP, while remaining comparable to more demanding methods such as MAD-ICP.

\begin{table}
\caption{Results on the MulRan~\cite{kim2020mulran} dataset. Best results are marked in \textbf{bold}.}
\label{tab:res_mulran}
\centering
\resizebox{\columnwidth}{!}{%
\begin{tabular}{l|rrrrr|rrrrr}
\specialrule{.17em}{.3em}{.3em} 
 & \multicolumn{5}{c|}{KITTI} & \multicolumn{5}{c}{$\text{RPE}_{100}$} \\
 \specialrule{.1em}{.3em}{.3em}  
\multirow{2}{*}{\centering} & \multicolumn{1}{c}{\raisebox{-.5\height}{Ours}} & \multicolumn{1}{c}{KISS-} & \multicolumn{1}{c}{MAD-} & \multicolumn{1}{c}{GenZ-} & \multicolumn{1}{c|}{KISS-} & \multicolumn{1}{c}{\raisebox{-.5\height}{Ours}} & \multicolumn{1}{c}{KISS-} & \multicolumn{1}{c}{MAD-}  & \multicolumn{1}{c}{GenZ-} & \multicolumn{1}{c}{KISS-}\\
&    & \multicolumn{1}{c}{ICP} & \multicolumn{1}{c}{ICP} & \multicolumn{1}{c}{ICP} & \multicolumn{1}{c|}{SLAM} &    & \multicolumn{1}{c}{ICP} & \multicolumn{1}{c}{ICP} & \multicolumn{1}{c}{ICP} & \multicolumn{1}{c}{SLAM}\\
\specialrule{.1em}{.3em}{.3em}  
DCC01 & \textbf{2.56} & 2.74 & 3.03 & 2.80 & 2.74 & \textbf{3.47} & 3.60 & 3.75 & 3.64 & 3.61 \\
DCC02 & \textbf{2.06} & 2.31 & 2.49 & 2.27 & 2.31 & \textbf{2.48} & 2.59 & 2.71 & 2.57 & 2.59 \\
DCC03 & \textbf{1.67} & 1.87 & 2.13 & 1.88 & 1.88 & \textbf{2.46} & 2.55 & 2.75 & 2.59 & 2.56 \\
KAIST01 & \textbf{1.97} & 2.09 & 2.58 & 2.26 & 2.09 & 2.31 & 2.28 & 2.58 & 2.45 & \textbf{2.28} \\
KAIST02 & \textbf{1.91} & 2.06 & 2.52 & 2.19 & 2.06 & \textbf{2.29} & 2.30 & 2.53 & 2.42 & 2.31 \\
KAIST03 & \textbf{2.28} & 2.43 & 2.92 & 2.54 & 2.43 & \textbf{2.57} & 2.60 & 2.85 & 2.69 & 2.60 \\
Riverside01 & \textbf{3.14} & 3.27 & 4.07 & 3.50 & 3.28 & \textbf{2.77} & 2.81 & 3.20 & 2.93 & 2.80 \\
Riverside02 & \textbf{3.06} & 3.12 & 3.69 & 3.25 & 3.12 & 2.73 & \textbf{2.59} & 2.90 & 2.71 & \textbf{2.59} \\
Riverside03 & \textbf{1.96} & 2.11 & 2.51 & 2.31 & 2.12 & 2.09 & 2.06 & 2.36 & 2.25 & \textbf{2.05} \\
Sejong01 & \textbf{3.53} & 4.44 & 4.97 & 4.77 & 4.77 & \textbf{2.66} & 2.94 & 3.12 & 3.14 & 3.11 \\
Sejong02 & \textbf{3.97} & 4.72 & 6.02 & 4.79 & 4.75 & \textbf{2.96} & 3.09 & 4.08 & 3.17 & 3.10 \\
Sejong03 & \textbf{4.30} & 4.89 & 5.69 & 5.43 & 4.90 & \textbf{3.20} & 3.26 & 3.75 & 3.85 & 3.26 \\
\specialrule{.1em}{.3em}{.3em} 
mean & \textbf{2.70} & 3.00 & 3.55 & 3.16 & 3.04 & \textbf{2.67} & 2.72 & 3.05 & 2.87 & 2.74 \\
\cmidrule{1-6}
FPS & 23.50 & \textbf{103.17} & 32.67 & 49.08 & 81.75 \\
\specialrule{.17em}{.3em}{.3em} 
\end{tabular}
}
\end{table}

\begin{table}
\caption{Results on the Odyssey~\cite{kurda2025odyssey} dataset. Best results are marked in \textbf{bold}.}
\label{tab:res_odyssey}
\centering
\resizebox{\columnwidth}{!}{%
\begin{tabular}{l|rrrrr|rrrrr}
\specialrule{.17em}{.3em}{.3em} 
 & \multicolumn{5}{c|}{KITTI} & \multicolumn{5}{c}{$\text{RPE}_{100}$} \\
 \specialrule{.1em}{.3em}{.3em}  
\multirow{2}{*}{\centering} & \multicolumn{1}{c}{\raisebox{-.5\height}{Ours}} & \multicolumn{1}{c}{KISS-} & \multicolumn{1}{c}{MAD-} & \multicolumn{1}{c}{GenZ-} & \multicolumn{1}{c|}{KISS-} & \multicolumn{1}{c}{\raisebox{-.5\height}{Ours}} & \multicolumn{1}{c}{KISS-} & \multicolumn{1}{c}{MAD-}  & \multicolumn{1}{c}{GenZ-} & \multicolumn{1}{c}{KISS-}\\
&    & \multicolumn{1}{c}{ICP} & \multicolumn{1}{c}{ICP} & \multicolumn{1}{c}{ICP} & \multicolumn{1}{c|}{SLAM} &    & \multicolumn{1}{c}{ICP} & \multicolumn{1}{c}{ICP} & \multicolumn{1}{c}{ICP} & \multicolumn{1}{c}{SLAM}\\
\specialrule{.1em}{.3em}{.3em}  
Beltway1 & \textbf{0.67} & 1.02 & 1.06 & 0.77 & 1.02 & 0.79 & 1.17 & 1.15 & \textbf{0.75} & 1.17 \\
Beltway2 & \textbf{0.65} & 1.02 & 1.01 & 0.77 & 1.02 & 0.75 & 1.15 & 1.15 & \textbf{0.71} & 1.15 \\
Beltway3 & \textbf{0.68} & 1.06 & 1.10 & 0.80 & 1.07 & 0.78 & 1.18 & 1.19 & \textbf{0.73} & 1.18 \\
CountryRoad1 & \textbf{0.76} & 1.13 & 1.03 & 0.95 & 1.13 & 0.74 & 1.14 & 1.19 & \textbf{0.68} & 1.14 \\
CountryRoad2 & \textbf{0.79} & 1.18 & 1.06 & 1.00 & 1.18 & 0.74 & 1.14 & 1.21 & \textbf{0.68} & 1.14 \\
CountryRoad3 & \textbf{0.80} & 1.19 & 1.04 & 1.04 & 1.19 & 0.74 & 1.15 & 1.21 & \textbf{0.67} & 1.15 \\
ForestRoad1 & \textbf{0.67} & 1.04 & 0.74 & 1.00 & 1.05 & \textbf{0.60} & 0.76 & 0.85 & 0.61 & 0.76 \\
ForestRoad2 & \textbf{0.69} & 1.14 & 0.72 & 1.09 & 1.14 & \textbf{0.55} & 0.75 & 0.79 & 0.57 & 0.75 \\
ForestRoad3 & \textbf{0.73} & 1.16 & 0.86 & 1.08 & 1.15 & \textbf{0.67} & 0.83 & 0.92 & 0.69 & 0.83 \\
Highway1 & \textbf{0.57} & 0.66 & 0.69 & 2.61 & 0.65 & \textbf{0.58} & 0.78 & 0.81 & 2.24 & 0.77 \\
Highway2 & \textbf{0.57} & 0.67 & 0.70 & 2.59 & 0.68 & \textbf{0.58} & 0.77 & 0.81 & 2.18 & 0.77 \\
Highway3 & \textbf{0.56} & 0.64 & 0.69 & 2.38 & 0.64 & \textbf{0.58} & 0.75 & 0.79 & 2.00 & 0.75 \\
HighwayTunnel1 & \textbf{0.62} & 0.78 & 0.67 & 0.90 & 0.78 & 0.57 & 0.70 & 0.75 & \textbf{0.50} & 0.70 \\
HighwayTunnel2 & \textbf{0.59} & 0.74 & 0.67 & 0.85 & 0.74 & 0.59 & 0.73 & 0.78 & \textbf{0.52} & 0.73 \\
HighwayTunnel3 & \textbf{0.67} & 0.90 & 0.69 & 0.92 & 0.87 & 0.61 & 0.83 & 0.77 & \textbf{0.50} & 0.80 \\
InnerCity1 & \textbf{0.56} & 0.68 & 0.72 & 0.70 & 0.67 & 0.59 & 0.76 & 0.76 & \textbf{0.50} & 0.76 \\
InnerCity2 & \textbf{0.60} & 0.73 & 0.75 & 0.67 & 0.73 & 0.77 & 0.96 & 0.92 & \textbf{0.66} & 0.96 \\
InnerCity3 & \textbf{0.53} & 0.66 & 0.67 & 0.64 & 0.67 & 0.61 & 0.87 & 0.83 & \textbf{0.52} & 0.87 \\
Marketplace1 & 0.56 & 0.87 & 0.88 & \textbf{0.54} & 0.87 & \textbf{0.63} & 0.94 & 0.93 & 0.64 & 0.94 \\
Marketplace2 & 0.58 & 0.87 & 0.95 & \textbf{0.56} & 0.88 & \textbf{0.67} & 1.03 & 1.02 & 0.68 & 1.03 \\
Marketplace3 & 0.64 & 0.99 & 1.05 & \textbf{0.61} & 1.00 & \textbf{0.74} & 1.17 & 1.17 & 0.76 & 1.18 \\
ParkingGarage1 & 0.55 & 0.64 & 0.62 & \textbf{0.34} & 0.64 & 0.77 & 0.93 & 0.89 & \textbf{0.51} & 0.93 \\
ParkingGarage2 & 0.55 & 0.69 & 0.63 & \textbf{0.39} & 0.69 & 0.81 & 1.00 & 0.90 & \textbf{0.60} & 1.00 \\
ParkingGarage3 & 0.61 & 0.63 & 0.60 & \textbf{0.36} & 0.63 & 0.91 & 0.94 & 0.89 & \textbf{0.58} & 0.94 \\
Suburb1 & \textbf{0.63} & 0.88 & 0.78 & 0.93 & 0.88 & \textbf{0.52} & 0.67 & 0.71 & \textbf{0.52} & 0.67 \\
Suburb2 & \textbf{0.64} & 0.90 & 0.86 & 0.92 & 0.90 & 0.53 & 0.71 & 0.74 & \textbf{0.46} & 0.71 \\
Suburb3 & \textbf{0.61} & 0.88 & 0.87 & 0.89 & 0.88 & \textbf{0.52} & 0.67 & 0.71 & \textbf{0.52} & 0.67 \\
Theater1 & \textbf{0.58} & 0.78 & 0.77 & 0.83 & 0.78 & 0.66 & 0.92 & 0.92 & \textbf{0.61} & 0.92 \\
Theater2 & \textbf{0.61} & 0.79 & 0.80 & 0.85 & 0.79 & 0.64 & 0.91 & 0.91 & \textbf{0.59} & 0.91 \\
Theater3 & \textbf{0.60} & 0.80 & 0.83 & 0.83 & 0.79 & 0.66 & 0.93 & 0.92 & \textbf{0.61} & 0.93 \\
UndergroundCarPark1 & \textbf{0.48} & 0.74 & 0.69 & 0.50 & 0.74 & \textbf{0.57} & 0.90 & 0.85 & 0.61 & 0.90 \\
UndergroundCarPark2 & \textbf{0.57} & 0.84 & 0.84 & 0.61 & 0.84 & \textbf{0.67} & 1.00 & 1.01 & 0.72 & 1.00 \\
UndergroundCarPark3 & \textbf{0.52} & 0.78 & 0.78 & \textbf{0.52} & 0.78 & \textbf{0.62} & 0.94 & 0.94 & 0.63 & 0.94 \\
\specialrule{.1em}{.3em}{.3em}
mean & \textbf{0.62}& 0.86 & 0.81 & 0.92 & 0.86 & \textbf{0.66} & 0.91 & 0.92 & 0.74 & 0.91 \\
\cmidrule{1-6}
FPS & 17.03 & \textbf{42.76} & 9.78 & 36.45 & 32.27 \\
\specialrule{.17em}{.3em}{.3em} 
\end{tabular}
}
\end{table}

\subsection{Comparison with FORM}
FORM~\cite{potokar2025form} is a recent lidar odometry method that, similar to our approach, enables retrospective map refinement through pose graph optimization. Different to the pose-pose constraints derived by our method, FORM constructs a densely connected graph where individual point-to-point and point-to-plane correspondences form constraints between the current frame and previous keyframes. Originally, FORM was evaluated primarily on handheld and slow-moving platforms, making a direct comparison with automotive-focused methods challenging. This domain difference is particularly evident in the original implementation's lack of a motion deskewing routine, a component essential for handling the higher velocities encountered in automotive datasets. To establish a feasible baseline for our experiments, we utilized the official implementation of FORM provided within the evalio\footnote{\url{https://github.com/contagon/evalio}} package and integrated a constant velocity motion deskewing routine, similar to the one used by our algorithm, to mitigate errors on faster sequences.
In our analysis, we observed that FORM exhibits limited applicability in the automotive domain. The method failed on 20 out of 56 sequences, spread over all datasets. After visual analysis of these sequences, we attribute this primarily to challenges posed by dynamic objects, particularly overtaking vehicles. FORM builds on the scanline-based feature extraction proposed by LOAM~\cite{zhang2014loam}, which, compared to the other methods, does not sample (approximately) uniformly in Cartesian space. Consequently, close objects such as overtaking cars contribute disproportionately many points, which can cause the method to diverge. While we attempted to mitigate these issues through sequence-specific parameter tuning, we were unable to identify a universal parameter set that ensured robust performance across all datasets and therefore excluded FORM from the quantitative analysis of the previous section.

\begin{table}[t]
\caption{Direct comparison between FORM and our method. Statistics are computed over all sequences successfully finished by FORM. Best results are marked in \textbf{bold}.}
\label{tab:5}
\centering
\begin{tabular}{l|rr|rr|rr}
\specialrule{.17em}{.3em}{.3em} 
& \multicolumn{2}{c|}{KITTI} & \multicolumn{2}{c|}{$\text{RPE}_{100}$} & \multicolumn{2}{c}{FPS} \\ 
 \specialrule{.1em}{.3em}{.3em}  
 & Ours & FORM & Ours & FORM & Ours & FORM \\
 \specialrule{.1em}{.3em}{.3em}  
 KITTI & \textbf{0.48} & 1.07 & \textbf{0.67} & 0.93 & \textbf{49.10} & 1.45 \\
 MulRan & \textbf{2.15} & 3.33 & \textbf{2.58} & 3.00 & \textbf{24.82} & 15.44 \\
 Odyssey & \textbf{0.6} & 0.78 & \textbf{0.69} & 0.76 & \textbf{15.40} & 1.19 \\ 
\specialrule{.17em}{.3em}{.3em} 
\end{tabular}
\end{table}

Table~\ref{tab:5} summarizes all sequences where FORM successfully completed the trajectory.
The results show that our method is generally more accurate and faster than FORM on the tested datasets. Closer analysis of the results by sequence revealed that the performance of FORM varies significantly with vehicle velocity.
The method performs notably worse than competing approaches on medium to high-velocity sequences, such as \texttt{KITTI 00} (Fig.~\ref{fig:3:kitti_00}) and this performance deficit persisted even in the total absence of dynamic vehicles, as seen in \texttt{ForestRoad1} (Fig.~\ref{fig:3:odyssey_forestroad1}). Conversely, FORM demonstrated very good accuracy on low-velocity sequences with minimal dynamic interference, such as \texttt{ParkingGarage} (Fig.~\ref{fig:3:odyssey_parkinggarege02}) or \texttt{UndergroundCarPark} sequences.
Our experiments suggest that, while capable of achieving high short-term consistency, FORM in its original implementation is not well suited for handling large dynamic objects and high velocities typically encountered in the automotive domain.

\section{Conclusion}\label{sec:conclusion}
In this work, we presented a lidar-only odometry method that leverages multiple independent registrations within a gated pose graph optimization framework to achieve high accuracy and robustness. By enabling retrospective map refinement through the use of a dynamic set of multiple overlapping submaps, our approach surpasses other state-of-the-art lidar odometry methods in terms of accuracy and robustness. Ablation studies confirm that both multi-submap registration and retrospective map refinement are critical for performance. The method achieves real-time operation while consistently outperforming competitors by 5–15\% in accuracy across multiple datasets. Preliminary results indicate strong potential for generalization beyond the automotive setting; however, further experiments are required for its validation. A limitation of our current approach is the reliance on fixed submap intervals, which leads to unnecessary redundancy in static regions as well as missing redundancy in changing environments. 
Here, future works should investigate the impact of a more sophisticated selection criterion based on, e.g, local motion or geometry to improve accuracy and reduce computational load.

\addtolength{\textheight}{-0cm}   

\bibliographystyle{IEEEtran}
\bibliography{./literature}

\end{document}